\documentclass[sigconf]{acmart}
\usepackage{subcaption}

\usepackage{pgfplots}
\usepackage{subcaption}
\usepackage{tikz}
\usepackage{float}
\usepackage{lipsum}
\usepackage{multirow}
\usepackage{wrapfig}
\usepackage{bm}

\AtBeginDocument{%
  }

\acmSubmissionID{8790}

\begin{document}

\title{MissDDIM: Deterministic and Efficient Conditional Diffusion for Tabular Data Imputation}

\author{Youran Zhou}
\email{echo.zhou@deakin.edu.au}
\affiliation{%
  \institution{Deakin University }
  \city{Geelong}
  \country{Australia}}

\author{Mohamed Reda Bouadjenek}
\affiliation{%
  \institution{Deakin University }
  \city{Geelong}
  \country{Australia}}

\author{Sunil Aryal}
\affiliation{%
  \institution{Deakin University }
  \city{Geelong}
  \country{Australia}}

\renewcommand{\shortauthors}{Zhou et al.}

\begin{abstract}
Diffusion models have recently emerged as powerful tools for missing data imputation by modeling the joint distribution of observed and unobserved variables. However, existing methods, typically based on stochastic denoising diffusion probabilistic models (DDPMs), suffer from high inference latency and variable outputs, limiting their applicability in real-world tabular settings. 
To address these deficiencies, we present in this paper \texttt{MissDDIM}, a conditional diffusion framework that adapts Denoising Diffusion Implicit Models (DDIM) for tabular imputation. While stochastic sampling enables diverse completions, it also introduces output variability that complicates downstream processing. 
\texttt{MissDDIM} replaces this with a deterministic, non-Markovian sampling path, yielding faster and more consistent imputations. To better leverage incomplete inputs during training, we introduce a self-masking strategy that dynamically constructs imputation targets from observed features—enabling robust conditioning without requiring fully observed data. Experiments on five benchmark datasets demonstrate that \texttt{MissDDIM} matches or exceeds the accuracy of state-of-the-art diffusion models, while significantly improving inference speed and stability. These results highlight the practical value of deterministic diffusion for real-world imputation tasks. Code will be available at: \url{https://anonymous.4open.science/r/MissDDIM-FE6E/}
\end{abstract}




\maketitle

\section{Introduction}

Missing data is a pervasive challenge in real-world applications such as healthcare~\cite{healthcare}, finance~\cite{financial}, recommendation systems~\cite{recommendation}, and sensor networks~\cite{sensor}. In tabular datasets, missing values degrade model performance and introduce bias or uncertainty in downstream analysis. Effective imputation—the task of estimating missing entries from observed data—is thus a critical step in many data-centric workflows. Traditional imputation methods such as mean/mode substitution~\cite{mean,mean2}, $k$-nearest neighbors~\cite{knn}, MICE~\cite{mice}, and MissForest~\cite{missforest} are efficient and stable, but often fail to capture feature dependencies and struggle under structured missingness. In contrast, deep generative models have shown great promise for modeling the joint distribution of observed and missing variables. Approaches based on GANs (e.g., GAIN~\cite{yoon2018gain}, MisGAN~\cite{misGAN}) and variational inference (e.g., MIWAE~\cite{mattei2019miwae}, HI-VAE~\cite{hivae}) have demonstrated stronger performance, while newer architectures like GRAPE~\cite{grape} and IGRM~\cite{igrm} incorporate iterative or graph-based interactions for greater expressiveness. More recently, diffusion models~\cite{dm1,dm2,dreview1} have emerged as powerful generative frameworks, particularly in vision and time series domains. Their gradual denoising process allows fine-grained, high-fidelity generation. In the tabular setting, several adaptations have been proposed: TabCSDI~\cite{tabcsdi} employs conditional score-based diffusion; MissDiff~\cite{missdiff} uses an unconditional formulation; and TabDDPM~\cite{tabddpm} extends diffusion to mixed-type data. Despite their modeling capacity, these methods face key limitations: (i) they rely on stochastic DDPM sampling, which incurs high inference latency and output variability; and (ii) many assume fully observed training data or lack robust conditioning on partial inputs—assumptions that rarely hold in practice~\cite{MLDL,Deeplearningisnotallyouneed}.

To address these challenges, we propose \textbf{MissDDIM}, the first framework to apply Denoising Diffusion Implicit Models (DDIM)~\cite{ddim} to imputation on incomplete tabular data. Unlike DDPMs, DDIM performs deterministic, non-Markovian sampling, enabling consistent outputs with significantly reduced inference cost. We further introduce a \textit{self-masking strategy} that dynamically creates training targets from partially observed data, allowing MissDDIM to learn directly from incomplete inputs. Through extensive experiments on five real-world datasets, we demonstrate that MissDDIM achieves competitive imputation accuracy while offering substantial improvements in inference speed and output stability. By bridging the gap between expressive generative modeling and efficient, deployment-friendly inference, MissDDIM provides a practical solution for real-world tabular imputation tasks.

\vspace{-0.2cm}

\section{The Proposed Method}
\subsection{Problem Formulation}

Let $\mathbf{X} \in \mathbb{R}^{n \times d}$ be a tabular dataset with $n$ samples and $d$ features, where each sample $\mathbf{x} \in \mathbb{R}^d$ is drawn from an unknown data distribution. Given a sample $\mathbf{x}_0$ that contains missing values, we aim to generate imputation targets $\mathbf{x}_0^{\text{mis}} \in \mathcal{X}^{\text{mis}}$ by exploiting the observed values $\mathbf{x}_0^{\text{obs}} \in \mathcal{X}^{\text{obs}}$, where $\mathcal{X}^{\text{mis}}$ and $\mathcal{X}^{\text{obs}}$ are subsets of the full feature space $\mathcal{X} = \mathbb{R}^d$.

\subsection{Denoising Diffusion Probabilistic Models}
\label{sec:background:ddpm}

Denoising Diffusion Probabilistic Models (DDPMs)~\citep{ddpm} model complex data distributions by simulating a Markov chain that gradually adds noise to the data (forward process) and then learns to reverse the corruption (reverse process). Let $\mathbf{x}_0 \in \mathcal{X}$ denote a data sample drawn from the unknown data distribution $q(\mathbf{x}_0)$. The forward process defines a sequence of latent variables $\mathbf{x}_1, \dots, \mathbf{x}_T$ in the same space $\mathcal{X}$, where $T$ is a predefined number of time steps. At each step $t \in \{1, \dots, T\}$, Gaussian noise is incrementally added to produce $\mathbf{x}_t$ from $\mathbf{x}_{t-1}$ according to:

\[
q(\mathbf{x}_t \mid \mathbf{x}_{t-1}) = \mathcal{N}(\mathbf{x}_t; \sqrt{1 - \beta_t} \mathbf{x}_{t-1}, \beta_t \mathbf{I}),
\]

where $\{\beta_t\}_{t=1}^T$ is a variance schedule controlling the amount of noise injected at each step. This leads to a closed-form marginal:
\begin{equation}
    q(\mathbf{x}_t \mid \mathbf{x}_0) = \mathcal{N}(\mathbf{x}_t; \sqrt{\alpha_t} \mathbf{x}_0,\ (1 - \alpha_t) \mathbf{I}),\text{ where }
    \alpha_t = \prod_{i=1}^t (1 - \beta_i).
\end{equation}

The reverse process is modeled as a denoising procedure learned via a parameterized distribution:
\begin{equation}
    p_\theta(\mathbf{x}_{t-1} \mid \mathbf{x}_t) = \mathcal{N}(\mathbf{x}_{t-1};\, \boldsymbol{\mu}_\theta(\mathbf{x}_t, t),\, \sigma_\theta^2(\mathbf{x}_t, t) \mathbf{I}),
\end{equation}
with $\boldsymbol{\mu}_\theta$ derived from a noise prediction network $\epsilon_\theta$:
\begin{equation}
    \boldsymbol{\mu}_\theta(\mathbf{x}_t, t) = \frac{1}{\alpha_t} \left( \mathbf{x}_t - \frac{\beta_t}{\sqrt{1 - \alpha_t}} \epsilon_\theta(\mathbf{x}_t, t) \right).
\end{equation}

While DDPMs yield high-quality generations, their reliance on long sampling chains (typically hundreds of steps) results in high inference latency, making them less practical for real-time imputation tasks that require rapid and stable outputs.

\begin{wrapfigure}{r}{0.48\linewidth}
    \vspace{-0.5em}
    \centering
    \includegraphics[width=\linewidth]{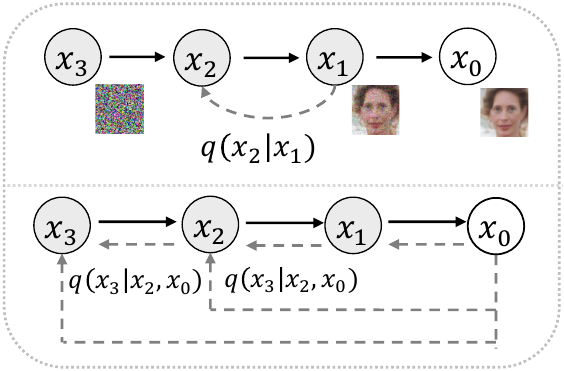}
    \caption{Different sampling processes: traditional stochastic diffusion (top) vs. deterministic non-Markovian inference (bottom).}
    \label{fig:fig_ddpm_ddim}
    \vspace{-1.5em}
\end{wrapfigure}

\subsection{MissDDIM: Efficient Imputation with Conditional DDIM}

While DDPMs have demonstrated strong generative capabilities, their sequential and stochastic nature leads to computationally expensive and inherently variable inference. These limitations pose practical challenges for missing value imputation in real-world applications, where stability and speed are crucial for downstream pipelines. To address this, we propose \texttt{MissDDIM}, a conditional diffusion model that adapts Denoising Diffusion Implicit Models (DDIM)~\citep{ddim} for efficient and deterministic imputation on tabular data. Unlike DDPMs, DDIM enables non-Markovian, parameter-free sampling trajectories, significantly reducing the number of inference steps while preserving generation quality (see Figure~\ref{fig:fig_ddpm_ddim}). Despite DDIM’s success in image synthesis, it has not yet been explored in the context of missing value imputation—particularly for tabular data, where heterogeneous feature types and partially observed inputs introduce unique challenges. \texttt{MissDDIM} bridges this gap by developing a conditional DDIM framework specifically tailored for imputation tasks.

\subsubsection{Conditional DDIM}

We aim to estimate the conditional distribution $p_\theta(\mathbf{x}_0^{\text{mis}} \mid \mathbf{x}_0^{\text{obs}})$, where the generative model focuses solely on missing components. To this end, we define a conditional noise prediction network
\[
\epsilon_\theta(\mathbf{x}_t^{\text{mis}}, t \mid \mathbf{x}_0^{\text{obs}})
\]
which predicts the noise applied to missing entries, given the observed context. This design explicitly conditions the reverse generation on known values at both training and inference stages, enabling targeted and stable imputation.

The reverse process is modified accordingly:
\begin{align}
p_\theta(\mathbf{x}_{0:T}^{\text{mis}} \mid \mathbf{x}_0^{\text{obs}}) 
&= p(\mathbf{x}_T^{\text{mis}}) \prod_{t=1}^T p_\theta(\mathbf{x}_{t-1}^{\text{mis}} \mid \mathbf{x}_t^{\text{mis}}, \mathbf{x}_0^{\text{obs}}), \\
p_\theta(\mathbf{x}_{t-1}^{\text{mis}} \mid \mathbf{x}_t^{\text{mis}}, \mathbf{x}_0^{\text{obs}})
&= \mathcal{N}\left(
\mathbf{x}_{t-1}^{\text{mis}};\;
\boldsymbol{\mu}_\theta(\mathbf{x}_t^{\text{mis}}, t \mid \mathbf{x}_0^{\text{obs}}),\;
\sigma_\theta^2(t)\, \mathbf{I}
\right),
\end{align}
where $\boldsymbol{\mu}_\theta$ is derived from the conditional noise prediction network~\citep{csdi} as:
\begin{equation}
\boldsymbol{\mu}_\theta(\mathbf{x}_t^{\text{mis}}, t \mid \mathbf{x}_0^{\text{obs}}) 
= \frac{1}{\sqrt{\alpha_t}} \left( 
\mathbf{x}_t^{\text{mis}} - \frac{\beta_t}{\sqrt{1 - \alpha_t}}\, \epsilon_\theta(\mathbf{x}_t^{\text{mis}}, t \mid \mathbf{x}_0^{\text{obs}})
\right).
\end{equation}
In standard DDPM-based samplers, the reverse process involves sampling from a Gaussian distribution with a learned mean and fixed variance. DDIM generalizes this by introducing a non-Markovian sampling schedule, where the level of stochasticity at each timestep is controlled by a noise parameter $\sigma_t$. When $\sigma_t > 0$, the process remains stochastic; when $\sigma_t = 0$, it becomes fully deterministic.

In \texttt{MissDDIM}, we adopt the deterministic variant by explicitly setting $\sigma_t = 0$ for all $t$, eliminating randomness in sampling and ensuring consistent outputs across runs—a critical property for reproducible imputation. The resulting update rule simplifies to:
\begin{equation}
\begin{aligned}
\mathbf{x}_{t-1}^{\text{mis}} =
&\ \sqrt{\alpha_{t-1}}\left(
\frac{\mathbf{x}_t^{\text{mis}} - \sqrt{1 - \alpha_t}\,\epsilon_\theta(\mathbf{x}_t^{\text{mis}}, t \mid \mathbf{x}_0^{\text{obs}})}
{\sqrt{\alpha_t}}
\right) \\
&\quad + \sqrt{1 - \alpha_{t-1}}\,\epsilon_\theta(\mathbf{x}_t^{\text{mis}}, t \mid \mathbf{x}_0^{\text{obs}}),
\end{aligned}
\label{eq:conditional_ddim}
\end{equation}

This deterministic formulation brings two key advantages:  
(i) it reduces inference latency by an order of magnitude due to fewer sampling steps;  
(ii) it ensures output consistency across runs, addressing the variability inherent in stochastic DDPM-based imputers.  
These properties make \texttt{MissDDIM} particularly suitable for latency-sensitive applications such as risk modeling, recommendation systems, and real-time analytics.

\subsubsection{Training Objective}
We adopt the standard DDPM/DDIM training strategy, adapted for conditional imputation. Given a sample with observed features $\mathbf{x}_0^{\text{obs}}$ and missing targets $\mathbf{x}_0^{\text{mis}}$, we corrupt the missing components using the forward diffusion process $\mathbf{x}_t^{\text{mis}} = \sqrt{\alpha_t},\mathbf{x}0^{\text{mis}} + \sqrt{1 - \alpha_t},\boldsymbol{\epsilon}$, where $\boldsymbol{\epsilon} \sim \mathcal{N}(\mathbf{0}, \mathbf{I})$. A conditional noise prediction network $\epsilon\theta(\mathbf{x}_t^{\text{mis}}, t \mid \mathbf{x}0^{\text{obs}})$ is trained to recover the injected noise, i.e., $\epsilon\theta(\mathbf{x}_t^{\text{mis}}, t \mid \mathbf{x}_0^{\text{obs}}) \approx \boldsymbol{\epsilon}$.
The model is optimized via a conditional denoising score matching loss:
\begin{equation}
\min_\theta \mathcal{L}(\theta) = \mathbb{E}_{\mathbf{x}_0,\, \boldsymbol{\epsilon},\, t} 
\left[ \left\| \boldsymbol{\epsilon} - \epsilon_\theta(\mathbf{x}_t^{\text{mis}}, t \mid \mathbf{x}_0^{\text{obs}}) \right\|_2^2 \right],
\end{equation}
where $\mathbf{x}_0 \sim q(\mathbf{x}_0)$ and $t \sim \text{Unif}(\{1, \dots, T\})$.

\subsubsection{Self-masking Strategy}

\begin{figure}[t]
    \centering
    \includegraphics[width=\linewidth]{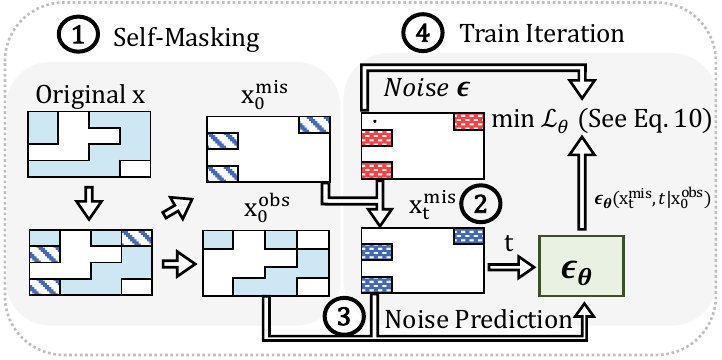}
    \caption{
    Illustration of the self-masking training strategy used in \texttt{MissDDIM}. 
    Observed values are randomly partitioned into conditional inputs and pseudo-targets during training.
    }\label{fig:fig_self_masking}
\end{figure}

Most generative imputation methods rely on prior imputers (e.g., mean filling) or externally provided masking vectors to handle missing inputs. In contrast, we adopt a \textbf{self-masking} strategy that enables the model to learn directly from partially observed data without auxiliary imputations (see Figure~\ref{fig:fig_self_masking}). Specifically, during training, a random subset of observed entries is masked and treated as targets, while the remaining observed values serve as conditional context.

\vspace{-0.2cm}
\subsection{Implementation}

Our implementation is based on the TabCSDI architecture~\citep{tabcsdi}, with key adaptations for non-sequential tabular data. Specifically, we remove the temporal transformer module—designed for time series—and retain only the feature-wise transformer encoder and residual MLP blocks to better suit static feature spaces.
Although \texttt{MissDDIM} builds upon TabCSDI, our DDIM-style sampler is model-agnostic. It depends only on the learned noise prediction network $\epsilon_\theta$, and is therefore compatible with any DDPM-based imputer, regardless of backbone. This makes \texttt{MissDDIM} a drop-in replacement for stochastic samplers, offering a general-purpose mechanism to accelerate and stabilize diffusion-based imputation pipelines.

\section{Experiments}

We evaluate \texttt{MissDDIM} from three perspectives: (i)~\textbf{Imputation accuracy}—how well the imputed values match ground truth; (ii)~\textbf{Sampling efficiency}—the trade-off between inference time and sampling steps; and (iii)~\textbf{Stability}—the consistency of results across runs. Together, these metrics demonstrate that \texttt{MissDDIM} achieves accurate, fast, and reliable imputation.

\subsection{Experimental Setup}
\subsubsection{Datasets}

We use five real-world datasets from the UCI Repository and Kaggle, covering both continuous-only data (\textit{Banknote}, \textit{California}, \textit{Letter}) and mixed-type data (\textit{Adult}, \textit{Student}). Missing values are simulated at four levels (10\%, 30\%, 50\%, 70\%). We evaluate both direct reconstruction error and downstream predictive performance.

\subsubsection{Baselines and Evaluation Protocol}
We compare \texttt{MissDDIM} against three categories of baselines: 
(i) statistical methods (\texttt{Mean/Mode}, \texttt{MICE}~\cite{mice}, \texttt{MissForest}~\cite{missforest}), 
(ii) deep generative models (\texttt{GAIN}~\cite{yoon2018gain}, \texttt{MIWAE}~\cite{mattei2019miwae}), and 
(iii) diffusion-based approaches (\texttt{CSDI}~\cite{csdi}, \texttt{TabCSDI}~\cite{tabcsdi}, \texttt{MissDiff}~\cite{missdiff}). All diffusion-based baselines use $T{=}100$ steps with 100 stochastic samples per instance (median-aggregated). \texttt{MissDDIM} yields deterministic imputations in a single forward pass.
All results are averaged over 5-fold cross-validation with 20\% of data held out for testing in each fold. Continuous features are standardized before imputation. Imputation accuracy is measured using RMSE averaged over missing entries. To evaluate downstream utility, we train an XGBoost classifier for classification tasks and XGBoost regressor for regression tasks on imputed data and report weighted F1-score or MAE respectively.

\begin{table*}[!t]
\centering
\caption{
Performance under 30\%, 50\%, and 70\% missingness across five datasets. 
Classification tasks are evaluated using weighted F1-score, while the \textit{Student} dataset (regression) uses MAE. 
Best results are shown in \textbf{bold}, and second-best results are \underline{underlined}. 
}
\label{tab:downstream_results}
\resizebox{\textwidth}{!}{
\begin{tabular}{l|ccc|ccc|ccc|ccc|ccc}
\hline
\textbf{Method} 
& \multicolumn{3}{c|}{\textbf{Adult}} 
& \multicolumn{3}{c|}{\textbf{Banknote}} 
& \multicolumn{3}{c|}{\textbf{California}} 
& \multicolumn{3}{c|}{\textbf{Letter}} 
& \multicolumn{3}{c}{\textbf{Student (MAE)}} \\
\hline
\textbf{Rate} & 30\% & 50\% & 70\% & 30\% & 50\% & 70\% & 30\% & 50\% & 70\% & 30\% & 50\% & 70\% & 30\% & 50\% & 70\% \\
\hline
Mean       & 0.4966 & 0.4750 & 0.4440 & 0.9000 & 0.7870 & 0.6703 & 0.8577 & 0.7595 & 0.6259 & 0.7960 & 0.6277 & 0.4046 & 2.2286 & 2.3608 & 2.5107 \\
MICE       & 0.6734 & 0.6447 & 0.6218 & 0.9416 & 0.8414 & 0.7512 & 0.7963 & 0.7373 & 0.6103 & 0.7862 & 0.7850 & 0.4370 & 2.0147 & 2.1097 & 2.4936 \\
MissForest & 0.6971 & 0.6517 & 0.6347 & 0.9517 & 0.8727 & 0.7821 & 0.8274 & 0.7556 & 0.6300 & 0.8323 & 0.8148 & 0.4524 & 1.7569 & 2.0165 & 1.9655 \\
\hline
GAIN       & 0.6048 & 0.5775 & 0.5441 & 0.9217 & 0.8725 & 0.7957 & 0.8291 & 0.7564 & 0.6094 & 0.8221 & 0.7567 & 0.4054 & 1.2574 & 1.4025 & 1.5993 \\
MIWAE      & 0.6941 & 0.6327 & 0.6014 & 0.9226 & 0.8937 & 0.8125 & 0.8531 & 0.6905 & 0.6261 & 0.8674 & 0.8186 & 0.4253 & 1.2347 & 1.4582 & 1.4614 \\\hline
CSDI     & 0.6827 & 0.6424 & 0.6218 & 0.9358 & 0.8867 & 0.8025 & 0.8446 & 0.8163 & 0.6291 & 0.8479 & 0.8090 & 0.4249 & 1.0257 & 1.2477 & 1.2098 \\

TabCSDI    & \underline{0.7214} & \underline{0.6851} & \textbf{0.6318} & \textbf{0.9527} & \underline{0.9014} & \underline{0.8657} & \underline{0.8886} & \underline{0.8386} & \underline{0.6289} & \underline{0.9049} & 0.8471 & 0.4467 & 0.8712 & 1.1371 & 1.1816 \\
MissDiff   & 0.7046 & 0.6711 & 0.6247 & 0.9416 & 0.8867 & 0.8237 & 0.8552 & 0.8253 & \textbf{0.6318} & 0.8503 & 0.8242 & \underline{0.4541} & \underline{0.8571} & \underline{1.1297} & \textbf{1.1575} \\
\hline
MissDDIM   & \textbf{0.7228} & \textbf{0.6724} & \textbf{0.6318} & \textbf{0.9527} & \textbf{0.9214} & \textbf{0.8772} & \textbf{0.8993} & \textbf{0.8608} & 0.6236 & \textbf{0.9117} & \textbf{0.8559} & \textbf{0.4581} & \textbf{0.8214} & \textbf{1.1143} & \underline{1.1713} \\
\hline
\end{tabular}
}
\end{table*}
\begin{table}[!t]
\centering
\caption{
Inference time and imputation accuracy (RMSE) comparison of generative imputation models. 
}
\label{tab:inference_summary}
\begin{tabular}{c|l|c|cc}
\hline
\#Samples & \textbf{Method} 
& Time (s) $\downarrow$ 
& RMSE $\downarrow$  \\
\hline
\multirow{2}{*}{100} 
& CSDI    & 820.73  & 0.3730 \\
& TabCSDI    & 794.21  & 0.3319  \\
& MissDiff   & 765.92  & 0.3163  \\
\hline
\multirow{3}{*}{1} 

& MissDDIM \textbf{(T = 100)} & \textbf{775.31}  & \textbf{0.3051}  \\
& MissDDIM \textbf{(T = 50)} & 385.12  & 0.3167  \\
& MissDDIM \textbf{(T = 20)} & \textbf{154.61}  & 0.3343 \\\hline
\end{tabular}
\end{table}
\section{Results and Analysis}

\subsection{Imputation Utility}

\begin{figure}[!t]
    \centering
    \includegraphics[width=\linewidth]{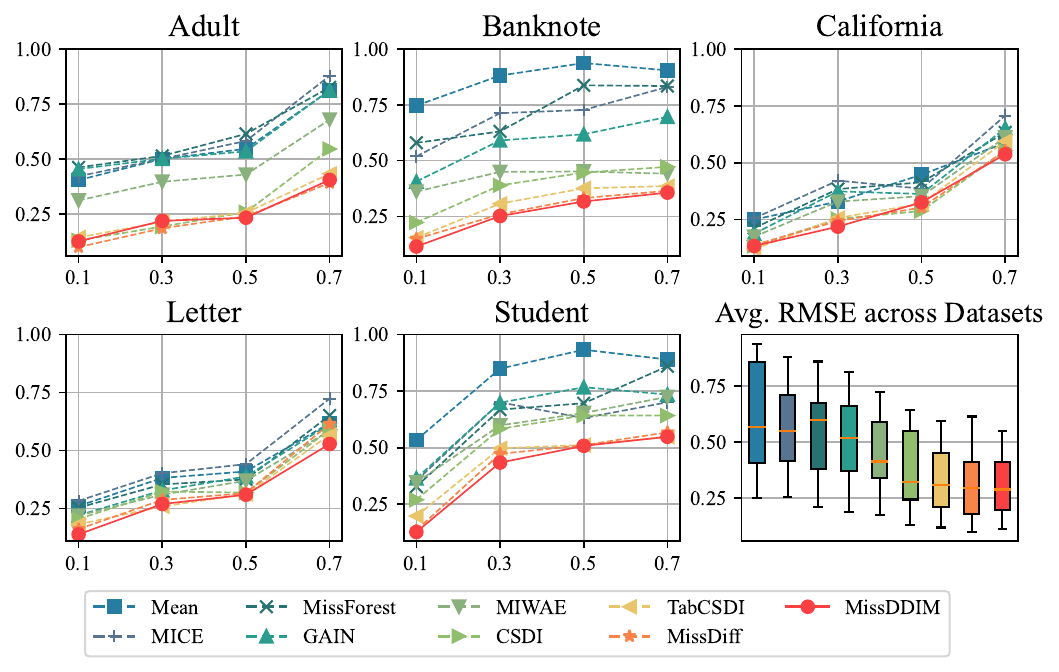}
    \caption{
RMSE across five benchmark datasets under varying missing rates. The final panel presents an aggregated summary of RMSE distributions using boxplots, combining results from all datasets and missingness levels.
    }
    \label{fig:rmse_mcar_grid}
    \vspace{-1em}
\end{figure}

Figure~\ref{fig:rmse_mcar_grid} summarises the imputation performance of competing methods under varying missing rates. As expected, the performance of all models generally degrades with increasing missingness. However, \texttt{MissDDIM} demonstrates relatively stable performance across different settings. The last subplot provides a boxplot that aggregates results across all datasets and missing rates, highlighting the superior stability of our method. Furthermore, Table~\ref{tab:downstream_results} shows that \texttt{MissDDIM} achieves strong performance in downstream predictive tasks, consistently outperforming or matching state-of-the-art baselines. Standard deviation regions are also reported to illustrate the robustness of our method.
\subsection{Sampling Time}
Existing diffusion-based models such as CSDI, TabCSDI and MissDiff typically generate 100 samples and take their median to produce stable imputations, which significantly increases inference time. In contrast, \texttt{MissDDIM} adopts a deterministic sampling process that requires only a single forward pass. Table~\ref{tab:inference_summary} reports the inference time and imputation accuracy under both standard (100-sample) and DDIM with single-sample settings for different T setted . For fairness, we evaluate all methods using the same batch size and computational environment. \texttt{MissDDIM} consistently demonstrates superior efficiency while maintaining competitive performance, confirming its practical advantage for real-time or large-scale imputation scenarios.

\subsection{Ablation Study}
\begin{figure}[!h]
\centering
\begin{subfigure}[b]{0.325\columnwidth}
    \centering
    \includegraphics[width=\linewidth]{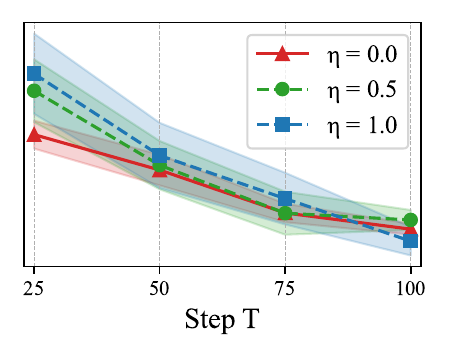}
        \vspace{-2em}
    \caption{Adult}
    \label{fig:ablation_sampling_adult}
\end{subfigure}
\hspace{-0.5em}
\begin{subfigure}[b]{0.325\columnwidth}
    \centering
    \includegraphics[width=\linewidth]{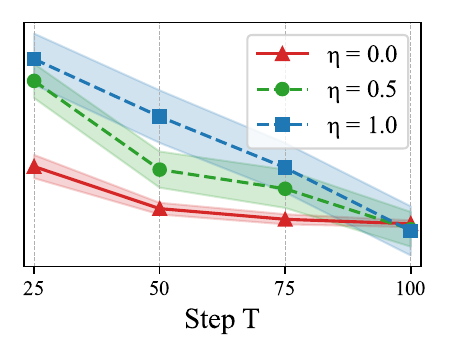}
            \vspace{-2em}
    \caption{Banknote}
    \label{fig:ablation_sampling_bank}
\end{subfigure}
\hspace{-0.5em}
\begin{subfigure}[b]{0.325\columnwidth}
    \centering
    \includegraphics[width=\linewidth]{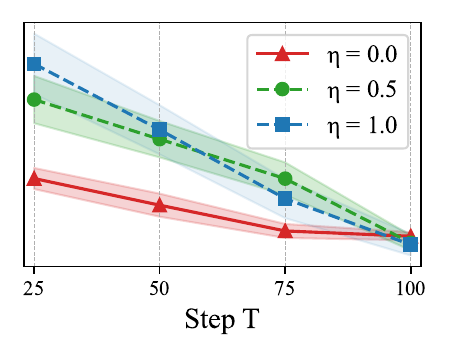}
            \vspace{-2em}
    \caption{California}
    \label{fig:ablation_sampling_cali}
\end{subfigure}
\vspace{-0.5em} 
\begin{subfigure}[b]{0.325\columnwidth}
    \centering
    \includegraphics[width=\linewidth]{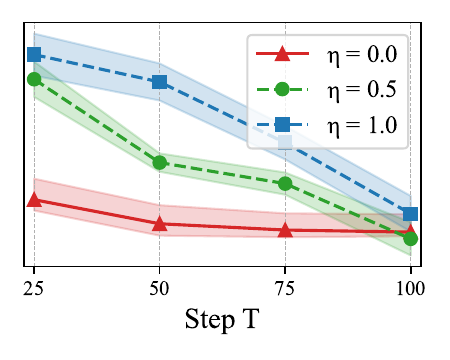}
            \vspace{-2em}
    \caption{Letter}
    \label{fig:ablation_sampling_letter}
\end{subfigure}
\hspace{-0.5em}
\begin{subfigure}[b]{0.325\columnwidth}
    \centering
    \includegraphics[width=\linewidth]{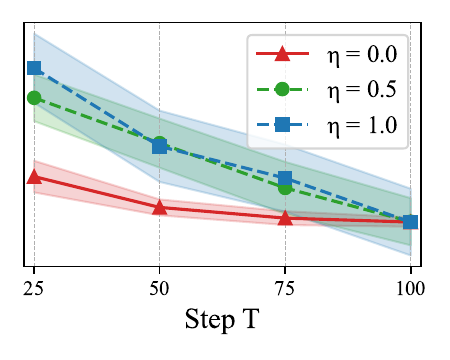}
            \vspace{-2em}
    \caption{Student}
    \label{fig:ablation_sampling_student}
\end{subfigure}
\hspace{-0.5em}
\begin{subfigure}[b]{0.325\columnwidth}
    \centering
    \includegraphics[width=\linewidth]{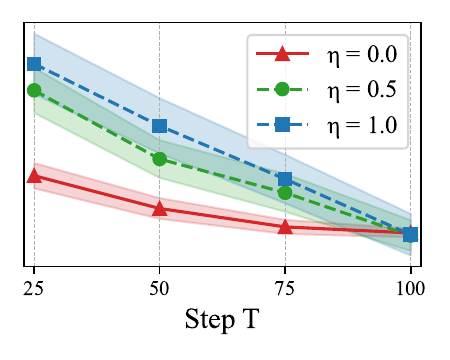}
            \vspace{-2em}
    \caption{Average}
    \label{fig:ablation_sampling_avg}
\end{subfigure}
\vspace{-0.5em}
\caption{RMSE Comparing different sampling strategies under varying $\eta$ values.}
\label{fig:ablation_sampling_combined}
\end{figure}

To evaluate the practical utility of \texttt{MissDDIM}, we assess its sampling efficiency and output stability under varying inference-time configurations. We control sampling stochasticity via the DDIM noise parameter $\eta \in \{0.0, 0.5, 1.0\}$, where $\eta{=}0$ yields fully deterministic trajectories (\texttt{MissDDIM}), and larger values inject increasing levels of noise, transitioning toward standard DDPM behavior. The variance at each timestep $\tau_i$ is defined as:
$
\sigma_{\tau_i}(\eta) = \eta \sqrt{\frac{1 - \alpha_{\tau_{i-1}}}{1 - \alpha_{\tau_i}}} \cdot \sqrt{1 - \frac{\alpha_{\tau_i}}{\alpha_{\tau_{i-1}}}},
$
which allows us to isolate the impact of stochasticity while keeping model parameters fixed. We also vary the number of sampling steps $T \in \{25, 50,75, 100\}$ to examine trade-offs between computational cost and imputation quality—reflecting latency-sensitive deployment scenarios. Table ~\ref{tab:eta_step_table} and Figure~\ref{fig:ablation_sampling_combined} report RMSE and its standard deviation across multiple runs. As expected, \texttt{MissDDIM} ($\eta{=}0$) produces deterministic outputs with low variance. At lower $T$, smaller $\eta$ values converge more rapidly and achieve competitive accuracy. As $T$ increases, stochastic methods may slightly improve performance but at the cost of higher variance. Overall, \texttt{MissDDIM} strikes an effective balance between inference speed, output stability, and imputation fidelity.

\section{Conclusion}

We proposed \texttt{MissDDIM}, the first imputation framework that adapts deterministic DDIM sampling to tabular data. By reformulating DDIM in a conditional setting, \texttt{MissDDIM} supports incomplete inputs natively and enables efficient, stable imputation without repeated sampling. Unlike existing diffusion-based approaches that rely on stochastic DDPM processes, our method achieves consistent outputs with significantly reduced inference time. Experiments across diverse datasets demonstrate that \texttt{MissDDIM} delivers competitive or superior accuracy, while offering practical advantages in latency-sensitive and deployment-oriented scenarios. In future work, we plan to extend \texttt{MissDDIM} to support more complex missing data mechanisms and explore strategies for improving robustness under  heterogeneous data type .

\begin{table}[!t]
\centering
\caption{Impact of sampling stochasticity ($\eta$) and number of sampling steps ($T$) on RMSE under 50\% missing rate. Lower values are better.}
\label{tab:eta_step_table}
\begin{tabular}{c|cc|cc|cc}
\hline
\multirow{2}{*}{$\eta$} & \multicolumn{2}{c|}{\textbf{Letter}} & \multicolumn{2}{c|}{\textbf{California}} & \multicolumn{2}{c}{\textbf{Average}} \\
\cline{2-7}
 & 25  & 100 & 25  & 100 & 25  & 100 \\
\hline
0.0 & \textbf{0.3214}  & 0.2254 & \textbf{0.5014}  & 0.3254 &\textbf{0.4977} & 0.3293\\
0.5 & 0.6780  & \textbf{0.2054} & 0.7421  & 0.3054 &0.7412  & \textbf{0.3167}\\
1.0 & 0.7510  & 0.2796 & 0.8510  & \textbf{0.2996} &0.8170 & 0.3343\\
\hline
\end{tabular}
\end{table}

\section*{Disclosure of AI Tools}
Generative AI tools were used to assist with language refinement in this manuscript. The authors manually revised all content.







\bibliographystyle{ACM-Reference-Format}
\bibliography{software}


\begin{thebibliography}{26}


\ifx \showCODEN    \undefined \def \showCODEN     #1{\unskip}     \fi
\ifx \showISBNx    \undefined \def \showISBNx     #1{\unskip}     \fi
\ifx \showISBNxiii \undefined \def \showISBNxiii  #1{\unskip}     \fi
\ifx \showISSN     \undefined \def \showISSN      #1{\unskip}     \fi
\ifx \showLCCN     \undefined \def \showLCCN      #1{\unskip}     \fi
\ifx \shownote     \undefined \def \shownote      #1{#1}          \fi
\ifx \showarticletitle \undefined \def \showarticletitle #1{#1}   \fi
\ifx \showURL      \undefined \def \showURL       {\relax}        \fi
\providecommand\bibfield[2]{#2}
\providecommand\bibinfo[2]{#2}
\providecommand\natexlab[1]{#1}
\providecommand\showeprint[2][]{arXiv:#2}

\bibitem[Al-taezi et~al\mbox{.}(2024)]%
        {misGAN}
\bibfield{author}{\bibinfo{person}{Mohammed~Ali Al-taezi}, \bibinfo{person}{Yu Wang}, \bibinfo{person}{Pengfei Zhu}, \bibinfo{person}{Qinghua Hu}, {and} \bibinfo{person}{Abdulrahman Al-Badwi}.} \bibinfo{year}{2024}\natexlab{}.
\newblock \showarticletitle{Improved generative adversarial network with deep metric learning for missing data imputation}.
\newblock \bibinfo{journal}{\emph{Neurocomputing}}  \bibinfo{volume}{570} (\bibinfo{year}{2024}), \bibinfo{pages}{127062}.
\newblock


\bibitem[Bryzgalova et~al\mbox{.}(2025)]%
        {financial}
\bibfield{author}{\bibinfo{person}{Svetlana Bryzgalova}, \bibinfo{person}{Sven Lerner}, \bibinfo{person}{Martin Lettau}, {and} \bibinfo{person}{Markus Pelger}.} \bibinfo{year}{2025}\natexlab{}.
\newblock \showarticletitle{Missing financial data}.
\newblock \bibinfo{journal}{\emph{The Review of Financial Studies}} \bibinfo{volume}{38}, \bibinfo{number}{3} (\bibinfo{year}{2025}), \bibinfo{pages}{803--882}.
\newblock


\bibitem[Croitoru et~al\mbox{.}(2023)]%
        {dreview1}
\bibfield{author}{\bibinfo{person}{Florinel-Alin Croitoru}, \bibinfo{person}{Vlad Hondru}, \bibinfo{person}{Radu~Tudor Ionescu}, {and} \bibinfo{person}{Mubarak Shah}.} \bibinfo{year}{2023}\natexlab{}.
\newblock \showarticletitle{Diffusion models in vision: A survey}.
\newblock \bibinfo{journal}{\emph{IEEE Transactions on Pattern Analysis and Machine Intelligence}} \bibinfo{volume}{45}, \bibinfo{number}{9} (\bibinfo{year}{2023}), \bibinfo{pages}{10850--10869}.
\newblock


\bibitem[Da’u and Salim(2020)]%
        {recommendation}
\bibfield{author}{\bibinfo{person}{Aminu Da’u} {and} \bibinfo{person}{Naomie Salim}.} \bibinfo{year}{2020}\natexlab{}.
\newblock \showarticletitle{Recommendation system based on deep learning methods: a systematic review and new directions}.
\newblock \bibinfo{journal}{\emph{Artificial Intelligence Review}} \bibinfo{volume}{53}, \bibinfo{number}{4} (\bibinfo{year}{2020}), \bibinfo{pages}{2709--2748}.
\newblock


\bibitem[Du et~al\mbox{.}(2020)]%
        {sensor}
\bibfield{author}{\bibinfo{person}{Jinghan Du}, \bibinfo{person}{Minghua Hu}, {and} \bibinfo{person}{Weining Zhang}.} \bibinfo{year}{2020}\natexlab{}.
\newblock \showarticletitle{Missing data problem in the monitoring system: A review}.
\newblock \bibinfo{journal}{\emph{IEEE Sensors Journal}} \bibinfo{volume}{20}, \bibinfo{number}{23} (\bibinfo{year}{2020}), \bibinfo{pages}{13984--13998}.
\newblock


\bibitem[Dutta and Sengupta(2016)]%
        {mice}
\bibfield{author}{\bibinfo{person}{Sulagna Dutta} {and} \bibinfo{person}{Pallav Sengupta}.} \bibinfo{year}{2016}\natexlab{}.
\newblock \showarticletitle{Men and mice: relating their ages}.
\newblock \bibinfo{journal}{\emph{Life sciences}}  \bibinfo{volume}{152} (\bibinfo{year}{2016}), \bibinfo{pages}{244--248}.
\newblock


\bibitem[Fei et~al\mbox{.}(2023)]%
        {dm2}
\bibfield{author}{\bibinfo{person}{Ben Fei}, \bibinfo{person}{Zhaoyang Lyu}, \bibinfo{person}{Liang Pan}, \bibinfo{person}{Junzhe Zhang}, \bibinfo{person}{Weidong Yang}, \bibinfo{person}{Tianyue Luo}, \bibinfo{person}{Bo Zhang}, {and} \bibinfo{person}{Bo Dai}.} \bibinfo{year}{2023}\natexlab{}.
\newblock \showarticletitle{Generative diffusion prior for unified image restoration and enhancement}. \bibinfo{publisher}{Proceedings of the IEEE/CVF conference on computer vision and pattern recognition}, \bibinfo{pages}{9935--9946}.
\newblock


\bibitem[Ho et~al\mbox{.}(2020)]%
        {ddpm}
\bibfield{author}{\bibinfo{person}{Jonathan Ho}, \bibinfo{person}{Ajay Jain}, {and} \bibinfo{person}{Pieter Abbeel}.} \bibinfo{year}{2020}\natexlab{}.
\newblock \showarticletitle{Denoising diffusion probabilistic models}. In \bibinfo{booktitle}{\emph{Proceedings of the 34th International Conference on Neural Information Processing Systems}} (Vancouver, BC, Canada) \emph{(\bibinfo{series}{NIPS '20})}. \bibinfo{publisher}{Curran Associates Inc.}, \bibinfo{address}{Red Hook, NY, USA}, Article \bibinfo{articleno}{574}, \bibinfo{numpages}{12}~pages.
\newblock
\showISBNx{9781713829546}


\bibitem[Houari et~al\mbox{.}(2014)]%
        {mean}
\bibfield{author}{\bibinfo{person}{Rima Houari}, \bibinfo{person}{Ahc{\`e}ne Bounceur}, \bibinfo{person}{A~Kamel Tari}, {and} \bibinfo{person}{M~Tahar Kecha}.} \bibinfo{year}{2014}\natexlab{}.
\newblock \showarticletitle{Handling missing data problems with sampling methods}. IEEE, \bibinfo{publisher}{2014 International conference on advanced networking distributed systems and applications}, \bibinfo{pages}{99--104}.
\newblock


\bibitem[Kotelnikov et~al\mbox{.}(2023)]%
        {tabddpm}
\bibfield{author}{\bibinfo{person}{Akim Kotelnikov}, \bibinfo{person}{Dmitry Baranchuk}, \bibinfo{person}{Ivan Rubachev}, {and} \bibinfo{person}{Artem Babenko}.} \bibinfo{year}{2023}\natexlab{}.
\newblock \showarticletitle{TabDDPM: modelling tabular data with diffusion models}. In \bibinfo{booktitle}{\emph{Proceedings of the 40th International Conference on Machine Learning}} (Honolulu, Hawaii, USA) \emph{(\bibinfo{series}{ICML'23})}. \bibinfo{publisher}{JMLR.org}, Article \bibinfo{articleno}{725}, \bibinfo{numpages}{16}~pages.
\newblock


\bibitem[Liu et~al\mbox{.}(2017)]%
        {healthcare}
\bibfield{author}{\bibinfo{person}{Caihua Liu}, \bibinfo{person}{Amir Talaei-Khoei}, \bibinfo{person}{Didar Zowghi}, {and} \bibinfo{person}{Jay Daniel}.} \bibinfo{year}{2017}\natexlab{}.
\newblock \showarticletitle{Data completeness in healthcare: a literature survey}.
\newblock \bibinfo{journal}{\emph{Pacific Asia Journal of the Association for Information Systems}} \bibinfo{volume}{9}, \bibinfo{number}{2} (\bibinfo{year}{2017}), \bibinfo{pages}{5}.
\newblock


\bibitem[Mattei and Frellsen(2019)]%
        {mattei2019miwae}
\bibfield{author}{\bibinfo{person}{Pierre-Alexandre Mattei} {and} \bibinfo{person}{Jes Frellsen}.} \bibinfo{year}{2019}\natexlab{}.
\newblock \showarticletitle{MIWAE: Deep generative modelling and imputation of incomplete data sets}. In \bibinfo{booktitle}{\emph{International conference on machine learning}}. PMLR, \bibinfo{pages}{4413--4423}.
\newblock


\bibitem[Nazabal et~al\mbox{.}(2020)]%
        {hivae}
\bibfield{author}{\bibinfo{person}{Alfredo Nazabal}, \bibinfo{person}{Pablo~M Olmos}, \bibinfo{person}{Zoubin Ghahramani}, {and} \bibinfo{person}{Isabel Valera}.} \bibinfo{year}{2020}\natexlab{}.
\newblock \showarticletitle{Handling incomplete heterogeneous data using vaes}.
\newblock \bibinfo{journal}{\emph{Pattern Recognition}}  \bibinfo{volume}{107} (\bibinfo{year}{2020}), \bibinfo{pages}{107501}.
\newblock


\bibitem[Ouyang et~al\mbox{.}(2023)]%
        {missdiff}
\bibfield{author}{\bibinfo{person}{Yidong Ouyang}, \bibinfo{person}{Liyan Xie}, \bibinfo{person}{Chongxuan Li}, {and} \bibinfo{person}{Guang Cheng}.} \bibinfo{year}{2023}\natexlab{}.
\newblock \bibinfo{title}{MissDiff: Training Diffusion Models on Tabular Data with Missing Values}.
\newblock
\showeprint[arxiv]{2307.00467}~[cs.LG]
\urldef\tempurl%
\url{https://arxiv.org/abs/2307.00467}
\showURL{%
\tempurl}


\bibitem[Shwartz-Ziv and Armon(2022)]%
        {Deeplearningisnotallyouneed}
\bibfield{author}{\bibinfo{person}{Ravid Shwartz-Ziv} {and} \bibinfo{person}{Amitai Armon}.} \bibinfo{year}{2022}\natexlab{}.
\newblock \showarticletitle{Tabular data: Deep learning is not all you need}.
\newblock \bibinfo{journal}{\emph{Information Fusion}}  \bibinfo{volume}{81} (\bibinfo{year}{2022}), \bibinfo{pages}{84--90}.
\newblock


\bibitem[Song et~al\mbox{.}(2021)]%
        {ddim}
\bibfield{author}{\bibinfo{person}{Jiaming Song}, \bibinfo{person}{Chenlin Meng}, {and} \bibinfo{person}{Stefano Ermon}.} \bibinfo{year}{2021}\natexlab{}.
\newblock \showarticletitle{Denoising Diffusion Implicit Models}. In \bibinfo{booktitle}{\emph{International Conference on Learning Representations}}.
\newblock
\urldef\tempurl%
\url{https://openreview.net/forum?id=St1giarCHLP}
\showURL{%
\tempurl}


\bibitem[Song and Shepperd(2007)]%
        {mean2}
\bibfield{author}{\bibinfo{person}{Qinbao Song} {and} \bibinfo{person}{Martin Shepperd}.} \bibinfo{year}{2007}\natexlab{}.
\newblock \showarticletitle{Missing data imputation techniques}.
\newblock \bibinfo{journal}{\emph{International journal of business intelligence and data mining}} \bibinfo{volume}{2}, \bibinfo{number}{3} (\bibinfo{year}{2007}), \bibinfo{pages}{261--291}.
\newblock


\bibitem[Stekhoven and Bühlmann(2011)]%
        {missforest}
\bibfield{author}{\bibinfo{person}{Daniel~J. Stekhoven} {and} \bibinfo{person}{Peter Bühlmann}.} \bibinfo{year}{2011}\natexlab{}.
\newblock \showarticletitle{MissForest—non-parametric missing value imputation for mixed-type data}.
\newblock \bibinfo{journal}{\emph{Bioinformatics}} \bibinfo{volume}{28}, \bibinfo{number}{1} (\bibinfo{date}{Oct.} \bibinfo{year}{2011}), \bibinfo{pages}{112–118}.
\newblock
\showISSN{1367-4803}
\href{https://doi.org/10.1093/bioinformatics/btr597}{doi:\nolinkurl{10.1093/bioinformatics/btr597}}


\bibitem[Tashiro et~al\mbox{.}(2021)]%
        {csdi}
\bibfield{author}{\bibinfo{person}{Yusuke Tashiro}, \bibinfo{person}{Jiaming Song}, \bibinfo{person}{Yang Song}, {and} \bibinfo{person}{Stefano Ermon}.} \bibinfo{year}{2021}\natexlab{}.
\newblock \showarticletitle{CSDI: conditional score-based diffusion models for probabilistic time series imputation}. In \bibinfo{booktitle}{\emph{Proceedings of the 35th International Conference on Neural Information Processing Systems}} \emph{(\bibinfo{series}{NIPS '21})}. \bibinfo{publisher}{Curran Associates Inc.}, \bibinfo{address}{Red Hook, NY, USA}, Article \bibinfo{articleno}{1900}, \bibinfo{numpages}{13}~pages.
\newblock
\showISBNx{9781713845393}


\bibitem[Xia et~al\mbox{.}(2023)]%
        {dm1}
\bibfield{author}{\bibinfo{person}{Bin Xia}, \bibinfo{person}{Yulun Zhang}, \bibinfo{person}{Shiyin Wang}, \bibinfo{person}{Yitong Wang}, \bibinfo{person}{Xinglong Wu}, \bibinfo{person}{Yapeng Tian}, \bibinfo{person}{Wenming Yang}, {and} \bibinfo{person}{Luc Van~Gool}.} \bibinfo{year}{2023}\natexlab{}.
\newblock \showarticletitle{Diffir: Efficient diffusion model for image restoration}. In \bibinfo{booktitle}{\emph{Proceedings of the IEEE/CVF International Conference on Computer Vision}}. \bibinfo{pages}{13095--13105}.
\newblock


\bibitem[Yoon et~al\mbox{.}(2018)]%
        {yoon2018gain}
\bibfield{author}{\bibinfo{person}{Jinsung Yoon}, \bibinfo{person}{James Jordon}, {and} \bibinfo{person}{Mihaela Schaar}.} \bibinfo{year}{2018}\natexlab{}.
\newblock \showarticletitle{Gain: Missing data imputation using generative adversarial nets}. In \bibinfo{booktitle}{\emph{International conference on machine learning}}. PMLR, \bibinfo{pages}{5689--5698}.
\newblock


\bibitem[You et~al\mbox{.}(2020)]%
        {grape}
\bibfield{author}{\bibinfo{person}{Jiaxuan You}, \bibinfo{person}{Xiaobai Ma}, \bibinfo{person}{Yi Ding}, \bibinfo{person}{Mykel~J Kochenderfer}, {and} \bibinfo{person}{Jure Leskovec}.} \bibinfo{year}{2020}\natexlab{}.
\newblock \showarticletitle{Handling missing data with graph representation learning}.
\newblock \bibinfo{journal}{\emph{Advances in Neural Information Processing Systems}}  \bibinfo{volume}{33} (\bibinfo{year}{2020}), \bibinfo{pages}{19075--19087}.
\newblock


\bibitem[Zhang(2012)]%
        {knn}
\bibfield{author}{\bibinfo{person}{Shichao Zhang}.} \bibinfo{year}{2012}\natexlab{}.
\newblock \showarticletitle{Nearest neighbor selection for iteratively kNN imputation}.
\newblock \bibinfo{journal}{\emph{Journal of Systems and Software}} \bibinfo{volume}{85}, \bibinfo{number}{11} (\bibinfo{year}{2012}), \bibinfo{pages}{2541--2552}.
\newblock


\bibitem[Zheng and Charoenphakdee(2022)]%
        {tabcsdi}
\bibfield{author}{\bibinfo{person}{Shuhan Zheng} {and} \bibinfo{person}{Nontawat Charoenphakdee}.} \bibinfo{year}{2022}\natexlab{}.
\newblock \showarticletitle{Diffusion models for missing value imputation in tabular data}. In \bibinfo{booktitle}{\emph{NeurIPS Table Representation Learning (TRL) Workshop}}.
\newblock


\bibitem[Zhong et~al\mbox{.}(2023)]%
        {igrm}
\bibfield{author}{\bibinfo{person}{Jiajun Zhong}, \bibinfo{person}{Ning Gui}, {and} \bibinfo{person}{Weiwei Ye}.} \bibinfo{year}{2023}\natexlab{}.
\newblock \showarticletitle{Data imputation with iterative graph reconstruction}. In \bibinfo{booktitle}{\emph{Proceedings of the AAAI Conference on Artificial Intelligence}}, Vol.~\bibinfo{volume}{37}. \bibinfo{pages}{11399--11407}.
\newblock


\bibitem[Zhou et~al\mbox{.}(2024)]%
        {MLDL}
\bibfield{author}{\bibinfo{person}{Youran Zhou}, \bibinfo{person}{Mohamed~Reda Bouadjenek}, {and} \bibinfo{person}{Sunil Aryal}.} \bibinfo{year}{2024}\natexlab{}.
\newblock \showarticletitle{Missing Data Imputation: Do Advanced ML/DL Techniques Outperform Traditional Approaches?}. In \bibinfo{booktitle}{\emph{Machine Learning and Knowledge Discovery in Databases. Applied Data Science Track}}, \bibfield{editor}{\bibinfo{person}{Albert Bifet}, \bibinfo{person}{Tomas Krilavi{\v{c}}ius}, \bibinfo{person}{Ioanna Miliou}, {and} \bibinfo{person}{Slawomir Nowaczyk}} (Eds.). \bibinfo{publisher}{Springer Nature Switzerland}, \bibinfo{address}{Cham}, \bibinfo{pages}{100--115}.
\newblock
\showISBNx{978-3-031-70381-2}


\end{thebibliography}





\end{document}